\documentclass[conference]{IEEEtran}
\IEEEoverridecommandlockouts
\usepackage{cite}
\usepackage{amsmath,amssymb,amsfonts}
\usepackage{algorithmic}
\usepackage{graphicx}
\usepackage{textcomp}
\usepackage{xcolor}
\usepackage{booktabs}
\usepackage{array}
\usepackage{url}
\def\BibTeX{{\rm B\kern-.05em{\sc i\kern-.025em b}\kern-.08em
    T\kern-.1667em\lower.7ex\hbox{E}\kern-.125emX}}

\begin{document}

\title{RL-Only Bootstrapping of OpenVLA-OFT for a Novel Cable-Driven Robot Embodiment}

\author{
\IEEEauthorblockN{Damir Nurtdinov}
\IEEEauthorblockA{%
Research Center of the Artificial Intelligence Institute \\
Innopolis University \\
Innopolis, Russia \\
\texttt{d.nurtdinov@innopolis.university}}
\and
\IEEEauthorblockN{Alexei Kornaev}
\IEEEauthorblockA{%
Center for top-educational programs in AI, Innopolis University \\
Innopolis, Russia\\
Moscow, Russia \\
\texttt{a.kornaev@innopolis.ru}}
\and
\IEEEauthorblockN{Alexander Maloletov}
\IEEEauthorblockA{%
Innopolis University \& Volgograd State Technical University \\
Innopolis \& Volgograd, Russia \\
\texttt{a.maloletov@innopolis.ru}}
}

\maketitle

\begin{abstract}
Adapting a pretrained vision-language-action (VLA) policy to a new robot usually assumes embodiment-specific demonstrations. This assumption is especially restrictive for custom robots whose morphology differs strongly from the manipulators seen in large robot datasets. We study a harder setting: zero-demo embodiment alignment of OpenVLA-OFT on a cable-driven parallel robot (CDPR) with a simple gripper and a previously unseen control interface. Instead of supervised fine-tuning, we use reinforcement learning in simulation with dense geometric rewards computed from simulator state. The training is performed in two stages: a PPO stage for directional motion primitives, followed by GRPO continuation from the PPO checkpoint with an expanded instruction space that includes object-conditioned commands. On the four shared directional instructions, the average held-out success rate improves from 34.25\% after PPO to 53.50\% after PPO$\rightarrow$GRPO, with especially large gains on \texttt{move left} and \texttt{move backward}. In the GRPO stage we additionally introduce \texttt{move to <object>} over eight target objects and obtain 39/400 = 9.75\% strict success, while qualitative rollouts frequently show correct target-directed approach behavior before late-stage instability. Compared with prior OpenVLA and OpenVLA-OFT results, which rely on demonstration datasets and mostly standard rigid-arm embodiments, our method uses no embodiment-specific dataset at all. The results do not yet establish robust manipulation, but they provide stronger evidence that RL-only bootstrapping can create the first usable language-conditioned controller for a genuinely novel embodiment.
\end{abstract}


\section{Introduction}
Pretrained vision-language-action (VLA) models are increasingly compelling as foundations for robot control because they combine language grounding, visual generalization, and parameter-efficient adaptation \cite{openvla,openvlaoft}. However, the standard adaptation recipe still assumes access to embodiment-specific demonstrations. OpenVLA studies fine-tuning on new Franka setups using 10--150 demonstrations per task \cite{openvla}, and OpenVLA-OFT improves this imitation-learning pipeline further with faster decoding and much higher task success \cite{openvlaoft,openvlaoftweb}. These are important advances, but they leave open a practically crucial question: what should a researcher do when the robot embodiment is new and no task dataset exists yet?

We study that setting directly. Specifically, we ask whether a pretrained OpenVLA-OFT policy can be aligned to a previously unseen cable-driven parallel robot (CDPR) through reinforcement learning alone, before collecting any embodiment-specific demonstrations. Because the robot is cable-driven and uses a minimal gripper, the problem is less about standard task transfer and more about establishing the first embodiment-compatible mapping between language, vision, and action.

Compared with our earlier PPO-only draft, this version strengthens the empirical case in three ways: it replaces the headline result with a cleaner four-direction benchmark that reaches 53.50\% average success after PPO$\rightarrow$GRPO continuation; it adds a new object-conditioned instruction family, \texttt{move to <object>}; and it clarifies the distinction from recent VLA-RL work by emphasizing dense simulator rewards rather than purely binary outcome rewards.

Our claim remains deliberately bounded: RL alone does not yet solve robust manipulation on a new embodiment, but it can produce non-trivial language-conditioned control without any embodiment-specific dataset and therefore serve as a useful first adaptation stage.

The contributions of this paper are:
\begin{itemize}
\item a zero-demo RL bootstrapping study for a novel cable-driven embodiment using OpenVLA-OFT as the policy backbone;
\item a dense-reward RL formulation that uses simulator geometry to provide embodiment-alignment signals before successful task completions are common;
\item an updated two-stage PPO$\rightarrow$GRPO result showing a directional-instruction average improvement from 34.25\% to 53.50\%; and
\item a comparative analysis against OpenVLA, OpenVLA-OFT, and recent VLA-RL post-training works that clarifies where the proposed setting is harder and where current headroom remains.
\end{itemize}

\section{Relation to Prior Work and Problem Setting}
OpenVLA is a 7B open-source VLA pretrained on 970k real-world robot demonstrations from Open X-Embodiment using 64 A100 GPUs for 15 days \cite{openvla,openvlaweb}. It supports strong zero-shot and fine-tuned performance, and for new Franka setups it is adapted with 10--150 demonstrations per task using LoRA updates to only 1.4\% of parameters \cite{openvla}.

OpenVLA-OFT strengthens this supervised pipeline with parallel chunk prediction and continuous L1 regression, improving LIBERO average success from 76.5\% to 97.1\% and increasing action-generation throughput by 26$\times$ \cite{openvlaoft,openvlaoftweb}. Its official recipe still assumes filtered successful demonstrations and uses 8 A100/H100 GPUs for 50K--150K steps \cite{openvlaoft,openvlaoftweb}.

Recent VLA-RL work also assumes a stronger starting point than ours. iRe-VLA alternates RL with supervised learning \cite{irevla}; RIPT-VLA uses sparse binary success rewards but still starts from at least one demonstration \cite{riptvla}; and SimpleVLA-RL emphasizes binary 0/1 outcome rewards with minimal reward engineering \cite{simplevlarl,simplevlarlcode}.

Our setting differs in one crucial respect: we assume \emph{no embodiment-specific dataset at all}. Dense simulator-side rewards are therefore important because early successful trajectories may be too rare for sparse-reward learning on a new embodiment.

\section{Embodiment-First Training Stack}
Our implementation\footnote{Git repository: \url{https://anonymous.4open.science/r/RL\_VLA\_Bootstrapping-E77F/README.md}.} combines MuJoCo embodiment modeling, reward definitions, scene generation, and OpenVLA-OFT RL fine-tuning in a single stack. The platform is a cable-driven parallel robot with a minimal gripper and a five-dimensional control interface: Cartesian end-effector motion in $x$, $y$, and $z$, yaw rotation, and gripper actuation. Low-level execution is handled by a PID controller in simulation.

The policy receives two RGB observations, an overview camera and a wrist-mounted camera, consistent with the OpenVLA-OFT multimodal interface \cite{openvlaoft}. To support systematic RL experiments, the stack includes:
\begin{itemize}
\item a MuJoCo embodiment specification and controller wrapper for the CDPR;
\item randomized scene generation built from YCB and LIBERO assets \cite{ycb,libero};
\item a shared action codec spanning the RL and downstream policy-execution paths; and
\item train/eval scripts for OpenVLA-OFT-based PPO and GRPO fine-tuning.
\end{itemize}

The key point is that the policy must align language and vision to a cable-driven actuation mechanism absent from the pretraining data.

\section{RL Bootstrapping Method}
\subsection{Stage 1: Directional PPO}
The first stage uses four primitive directional instructions:
\begin{quote}
\small \texttt{move left}, \texttt{move right}, \texttt{move forward}, \texttt{move backward}.
\end{quote}
These commands teach the policy how instruction semantics map into the CDPR action space before object-conditioned behaviors are introduced.

\subsection{Stage 2: GRPO Continuation with Object-Conditioned Language}
The second stage continues training from the PPO checkpoint using GRPO and expands the instruction space to include
\begin{quote}
\small \texttt{move to <object>},
\end{quote}
where the target object is sampled from eight categories: apple, baseball, bowl, cup, mug, peach, pear, and plate. This stage tests whether the policy can move beyond directional grounding while preserving the gains from Stage 1.

\subsection{Dense Reward Design}
The central algorithmic choice is to use dense rewards computed from simulator geometry. Let $d_t$ denote the current distance between the end effector and the instruction-dependent target (a directional target region or an object target), and let $a_t$ denote the policy action. We optimize a progress-based reward of the form
\begin{equation}
r_t = w_p (d_{t-1} - d_t) + b_s \,\mathbb{I}[\mathrm{succ}_t] - w_a P(a_t),
\end{equation}
where $\mathrm{succ}_t$ is a task-specific binary success indicator and $P(a_t)$ penalizes near-saturated non-gripper actions. The reward therefore encourages progress to the target, gives a success bonus, and discourages unstable control saturation.

This choice matters because recent VLA-RL work often emphasizes sparse or binary outcome rewards \cite{riptvla,simplevlarl,simplevlarlcode}, whereas our no-demo cable-driven setting requires a shaped signal before consistently successful trajectories exist.

\subsection{Training Protocol}
The policy backbone is OpenVLA-OFT initialized from the public \texttt{openvla/openvla-7b} checkpoint \cite{openvlaoft}. We train adapters and action-head parameters with two image inputs and an 8-step action chunk. Stage 1 uses PPO for approximately 175 hours on two NVIDIA A40 GPUs; Stage 2 continues with GRPO for another 170 hours on the same hardware. The full RL budget is therefore about 345 hours with zero embodiment-specific demonstrations.

\section{Evaluation Protocol}
We evaluate on held-out randomized scenes using the repository validator. Each directional instruction is tested for 100 rollouts, and the object-conditioned instruction family for 400 rollouts in total. Success rate is the main quantitative metric.

For the object-conditioned stage, we also report qualitative evidence because the strict metric undercounts episodes in which the end effector approaches the correct target but drifts away late in the rollout.

\section{Results}
\subsection{PPO to GRPO Improvement}
Table~\ref{tab:ppo_grpo} summarizes the main quantitative result. On the four directional instructions shared across both stages, the mean success rate increases from 34.25\% after PPO to 53.50\% after PPO$\rightarrow$GRPO. The largest gains are on \texttt{move left} (+35 percentage points) and \texttt{move backward} (+33 percentage points), while \texttt{move forward} remains at 62\%.

\begin{table}[t]
\caption{Validation comparison between the PPO checkpoint and the continued PPO$\rightarrow$GRPO model on the four shared directional instructions.}
\label{tab:ppo_grpo}
\centering
\footnotesize
\begin{tabular}{lccc}
\toprule
Instruction & PPO (\%) & PPO$\rightarrow$GRPO (\%) & $\Delta$ (pp) \\
\midrule
Move left & 17.00 & 52.00 & +35.00 \\
Move right & 43.00 & 52.00 & +9.00 \\
Move forward & 62.00 & 62.00 & +0.00 \\
Move backward & 15.00 & 48.00 & +33.00 \\
\midrule
Mean (4 directions) & 34.25 & 53.50 & +19.25 \\
\bottomrule
\end{tabular}
\end{table}

This result addresses the main weakness of the earlier PPO-only version: continued RL materially improves a cleaner four-direction benchmark without collapsing the best-performing instruction.

\subsection{Object-Conditioned Evidence}
The expanded GRPO stage introduces \texttt{move to <object>}, evaluated across eight target objects. The strict validator reports 39 successes out of 400 rollouts, i.e., 9.75\%. Although this remains low, many validation episodes show the policy approaching the correct object before late-stage instability causes failure.

\begin{figure*}[t]
\centering
\begin{tabular}{cccc}
\includegraphics[width=0.22\textwidth]{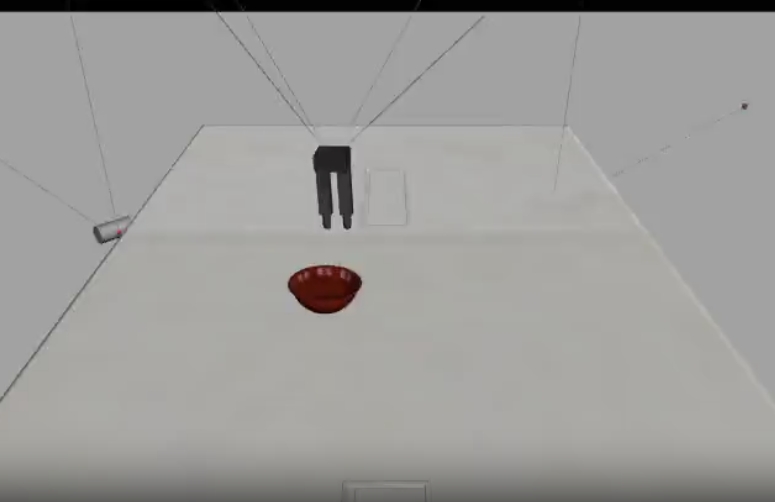} &
\includegraphics[width=0.22\textwidth]{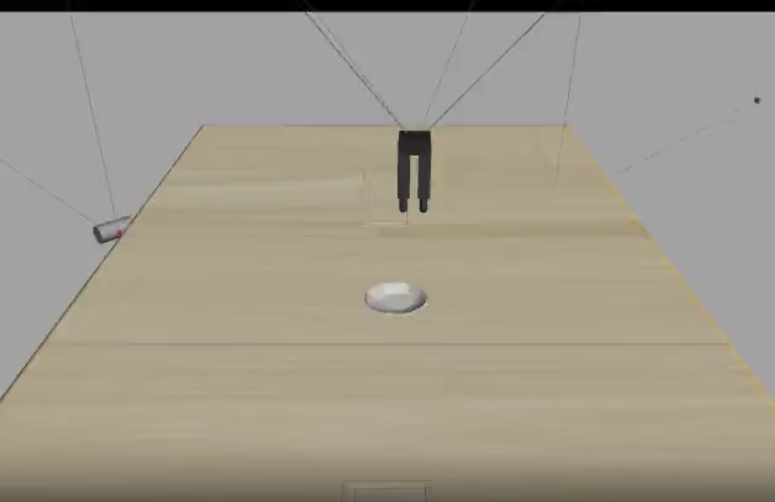} &
\includegraphics[width=0.22\textwidth]{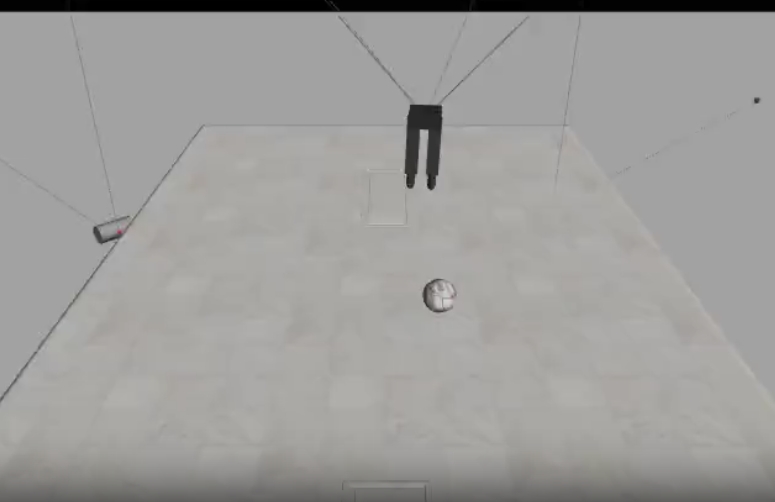} &
\includegraphics[width=0.22\textwidth]{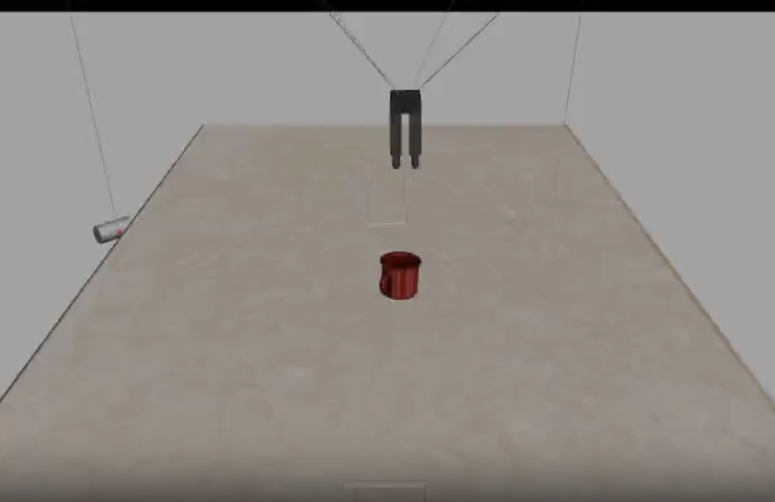} \\
(a) Bowl & (b) Plate & (c) Baseball & (d) Mug \\
\includegraphics[width=0.22\textwidth]{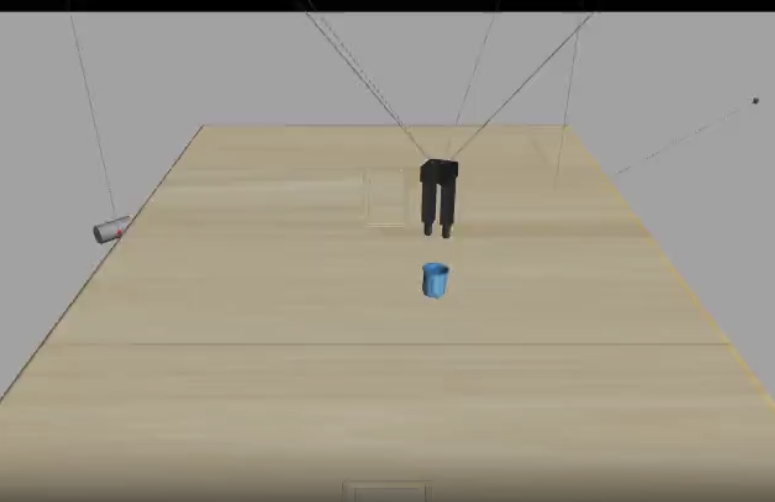} &
\includegraphics[width=0.22\textwidth]{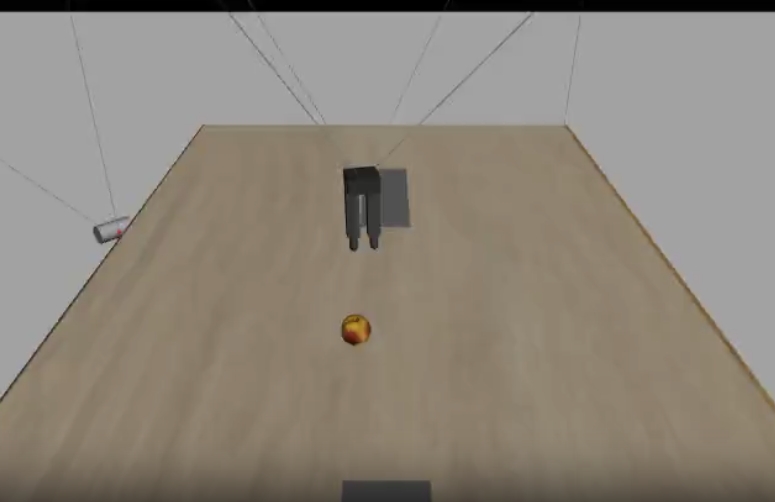} &
\includegraphics[width=0.22\textwidth]{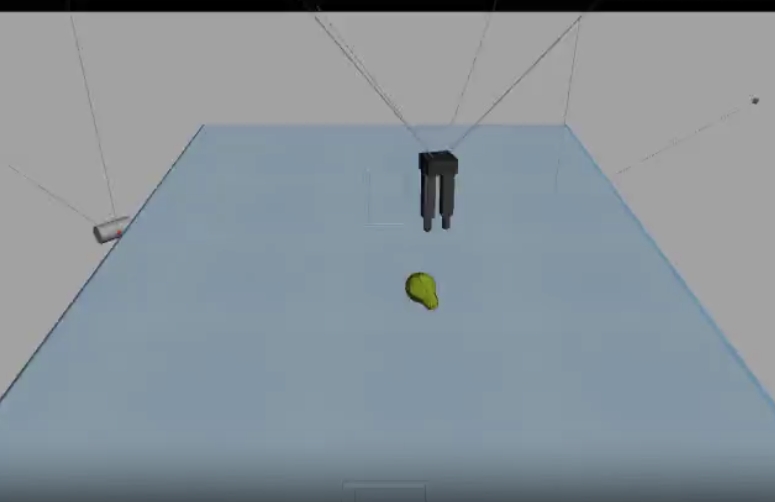} &
\includegraphics[width=0.22\textwidth]{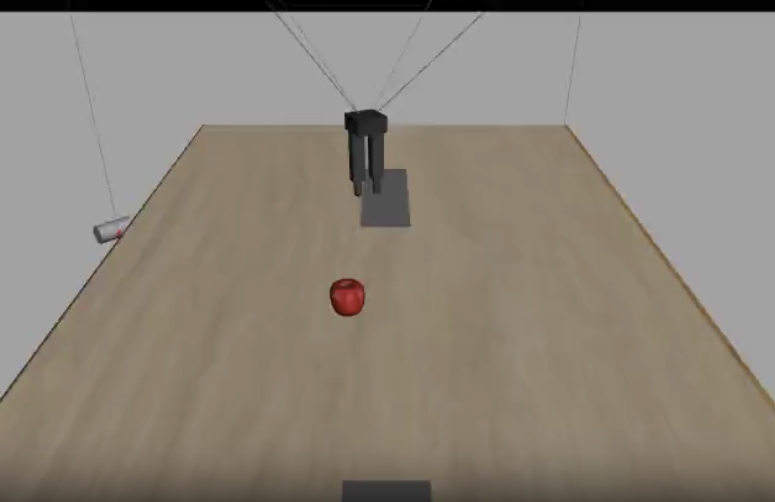} \\
(e) Cup & (f) Peach & (g) Pear & (h) Apple
\end{tabular}
\caption{Representative validation rollouts for \texttt{move to <object>}. Despite a strict success rate of 39/400 = 9.75\%, the policy often approaches the correct target before late-rollout instability causes failure.}
\label{fig:move_to_object}
\end{figure*}

Figure~\ref{fig:move_to_object} shows that the current failure mode is usually late-stage instability rather than missing object grounding from scratch.

\subsection{Why Dense Reward Matters in This Setting}
Dense reward is central to the scientific argument. Sparse binary post-training methods such as RIPT-VLA and SimpleVLA-RL improve VLAs when at least a small supervised starting point already exists \cite{riptvla,simplevlarl}. Our harsher zero-demo setting instead needs dense reward shaping to make early embodiment alignment tractable.

\section{Discussion}
RL-only embodiment bootstrapping is now supported by substantially stronger evidence than in the earlier PPO-only draft: the directional benchmark reaches a 53.50\% mean after PPO$\rightarrow$GRPO continuation, and the policy additionally exhibits object-conditioned behavior on eight categories without any embodiment-specific dataset.

The result should still be interpreted carefully. The experiments remain in simulation, the object-conditioned task is far from robust under strict evaluation, and the success rates are not directly comparable to the best OpenVLA/OFT numbers because those works rely on different tasks and substantial supervised data. The evidence therefore supports a staged methodology:
\begin{enumerate}
\item use RL with dense simulator rewards to obtain the first embodiment-aligned language-conditioned controller;
\item use that controller to reduce the cost of collecting downstream data or to initialize later imitation learning; and
\item continue post-training with richer instructions, stricter evaluation, and eventually real-robot transfer.
\end{enumerate}

The observed ``pushcut''-style shortcut behavior, similar to the phenomenon reported in SimpleVLA-RL \cite{simplevlarl}, is also useful scientifically: it indicates genuine reward-driven adaptation and motivates improved reward design near the goal.

\section{Conclusion}
We presented a revised study of RL-only embodiment alignment for OpenVLA-OFT on a novel cable-driven robot. Unlike standard VLA adaptation pipelines, the proposed method uses no embodiment-specific demonstrations. Instead, it relies on dense simulator rewards and a two-stage PPO$\rightarrow$GRPO curriculum. The updated experiments improve the four-direction mean success rate from 34.25\% to 53.50\% and extend the instruction space to object-conditioned commands, where the policy already shows clear target-directed approach behavior despite a still-low strict success rate. Taken together, these results strengthen the original thesis: RL can serve as a practical bootstrapping stage for bringing a pretrained VLA model onto a genuinely new embodiment before any task dataset exists.

\end{document}